\PassOptionsToPackage{prologue,dvipsnames}{xcolor}
\documentclass[letterpaper, 10 pt, journal, twoside]{IEEEtran}
\usepackage{amsmath,amsfonts}
\usepackage{algorithmic}
\usepackage{algorithm}
\usepackage{array}
\usepackage[caption=false,font=normalsize,labelfont=sf,textfont=sf]{subfig}
\usepackage{textcomp}
\usepackage{stfloats}
\usepackage{url}
\usepackage{verbatim}
\usepackage{graphicx}
\usepackage{booktabs}
\usepackage{multirow}
\usepackage{pifont}
\usepackage{cite}
\usepackage{xcolor}

\usepackage{amsmath,amsfonts,bm}

\def\eqref#1{equation~\ref{#1}}

\def\1{\bm{1}}

\def\vo{{\bm{o}}}
\def\vp{{\bm{p}}}

\def\mF{{\bm{F}}}

\def\mI{{\bm{I}}}

\def\mK{{\bm{K}}}

\def\mT{{\bm{T}}}

\DeclareMathAlphabet{\mathsfit}{\encodingdefault}{\sfdefault}{m}{sl}
\SetMathAlphabet{\mathsfit}{bold}{\encodingdefault}{\sfdefault}{bx}{n}

\newcommand{\cmark}{\ding{51}}%
\newcommand{\xmark}{\ding{55}}%

\begin{document}

\title{RoDyn-SLAM: Robust Dynamic Dense RGB-D SLAM with Neural Radiance Fields}

\author{Haochen Jiang, Yueming Xu, Kejie Li, Jianfeng Feng, Li Zhang
\thanks{Manuscript received February 28, 2024; Revised May 29, 2024; Accepted June 26, 2024. Haochen Jiang, Yueming Xu contributed equally to this work. \textit{(Corresponding author: Li Zhang (e-mail: lizhangfd@fudan.edu.cn) with School of Data Science, Fudan University)}}
\thanks{Haochen Jiang is with the School of Data Science, Fudan University, Shanghai 200433, China. E-mail: jhch1995@mail.ustc.edu.cn. Yueming Xu and Jianfeng Feng are with the Institute of Science and Technology for Brain-Inspired Intelligence, Fudan University, Shanghai 200433, China. E-mails: xuyueming21@m.fudan.edu.cn, jffeng@fudan.edu.cn. Kejie Li is with ByteDance, Seattle, USA. E-mail: kejie.li@outlook.com.}
\thanks{$^1$https://github.com/fudan-zvg/Rodyn-SLAM}
}

\markboth{IEEE Robotics and Automation Letters. Preprint Version. Accepted June, 2024}%
{Jiang \MakeLowercase{\textit{et al.}}: RoDyn-SLAM: Robust Dynamic Dense RGB-D SLAM with Neural Radiance Fields} 


\maketitle

\begin{abstract}
Leveraging neural implicit representation to conduct dense RGB-D SLAM has been studied in recent years. However, this approach relies on a static environment assumption and does not work robustly within a dynamic environment due to the inconsistent observation of geometry and photometry. 
To address the challenges presented in dynamic environments, we propose a novel dynamic SLAM framework with neural radiance field. Specifically, we introduce a motion mask generation method to filter out the invalid sampled rays. This design effectively fuses the optical flow mask and semantic mask to enhance the precision of motion mask. To further improve the accuracy of pose estimation, we have designed a divide-and-conquer pose optimization algorithm that distinguishes between keyframes and non-keyframes. The proposed edge warp loss can effectively enhance the geometry constraints between adjacent frames. Extensive experiments are conducted on the two challenging datasets, and the results show that RoDyn-SLAM achieves state-of-the-art performance among recent neural RGB-D methods in both accuracy and robustness. Our implementation of the Rodyn-SLAM will be open-sourced to benefit the community$^1$.
\end{abstract}

\begin{IEEEkeywords}
Deep Learning Methods, NeRF, RGB-D SLAM, Dynamic Scene, Pose Estimation.
\end{IEEEkeywords}

\section{Introduction}
\IEEEPARstart{D}{ense} visual simultaneous localization and mapping (SLAM) is a fundamental task in 3D computer vision and robotics, which has been widely used in various forms in fields such as service robotics, autonomous driving, and augmented/virtual reality (AR/VR). It is defined as reconstructing a dense 3D map in an unknown environment while simultaneously estimating the camera pose, which is regarded as the key to achieving autonomous navigation for robots~\cite{leonard1991simultaneous}. However, the majority of methods assume a static environment, limiting the applicability of this technology to more practical scenarios.
Thus, it becomes a challenging problem that how the SLAM system can mitigate the interference caused by dynamic objects.  

Traditional visual SLAM methods using semantic segmentation prior~\cite{ds_slam,dynaslam,dynamicslam,vdoslam}, optical flow motion~\cite{sun2018motion,cheng2019improving,flowfusion} or re-sampling and
residual optimization strategies~\cite{orbslam2,orbslam3,refusion} to remove the outliers under dynamic environments, which can improve the accuracy and robustness of pose estimation. However, re-sampling and optimization methods can only handle small-scale motions and often fail when encountering large-scale continuous object movements. Moreover, semantic priors are specific to particular categories and can not represent the real motion state of the observation object. The above learning-based methods often exhibit a domain gap when applied in real-world environments, leading to the introduction of prediction errors.     

Recently, dense visual SLAM with neural implicit representation has gained more attention and popularity. This novel map representation is more compact, continuous, efficient, and able to be optimized with differentiable rendering, which has the potential to benefit applications like navigation, planning, and reconstruction. Moreover, the neural scene representations have attractive properties for mapping, including improving noise and outlier handling, geometry estimation capabilities for unobserved scene parts, high-fidelity reconstructions with reduced memory usage, and the ability to generate high-quality static background images from novel views. Existing methods like iMap~\cite{imap} and NICE-SLAM~\cite{nice_slam} respectively leverage single MLP and hierarchical feature grids to achieve a consistent geometry representation. However, these methods have limited capacity to capture intricate geometric details. Recent works such as Co-SLAM~\cite{co-slam} and ESLAM~\cite{eslam} explore hash encoding or tri-plane representation strategy to enhance the capability of scene representation and the system's execution efficiency. However, all these above-mentioned methods do not perform well in dynamic scenes. The robustness of these systems significantly decreases, even leading to tracking failures when dynamic objects appear in the environment.

To tackle these problems, we propose a novel NeRF-based RGB-D SLAM that can reliably track camera motion in indoor dynamic environments.
One of the key elements to improve the robustness of pose estimation is the motion mask generation algorithm that filters out the sampled rays located in invalid regions. 
By incrementally fusing the optical flow mask~\cite{gma}, the semantic segmentation mask~\cite{oneformer} can become more precise to reflect the true motion state of objects.
To further improve the accuracy of pose estimation, we design a divide-and-conquer pose optimization algorithm for keyframes and non-keyframes.
While an efficient edge warp loss is used to track camera motions for all keyframes and non-keyframes w.r.t. adjacent frames,
only keyframes are further jointly optimized via rendering loss in the global bundle adjustment (GBA). 

In summary, our \textbf{contributions} are summarized as follows:
\begin{enumerate}
    \item {To the best of our knowledge, this is the first dynamic neural RGB-D SLAM with joint robust pose estimation and dense reconstruction.} 
    \item {In response to the issue of inaccurate semantic priors, we propose a motion mask generation strategy fusing spatial-temporal consistent optical flow masks to improve the robustness of camera pose estimation and quality of static scene reconstruction.}
    \item {Instead of a single frame tracking method, we design a novel mixture pose optimization algorithm utilizing an edge warp loss to enhance the geometry consistency in the non-keyframe tracking stage.}
    \item {We evaluate our method on two challenging dynamic datasets to demonstrate the state-of-the-art performance of our method in comparison to existing NeRF-based RGB-D SLAM approaches.}
\end{enumerate}

\section{Related work}
\subsection{Conventional visual SLAM with dynamic objects filter}
Dynamic object filtering aims to reconstruct the static scene and enhance the robustness of pose estimation. Prior methods can be categorized into two groups: the first one utilizes the re-sampling and residual optimization strategies to remove the outliers~\cite{orbslam2,orbslam3,refusion}. 
However, these methods can only handle small-scale motions and often fail when encountering large-scale continuous object movements. 
The second group employs the additional prior knowledge, such as semantic segmentation prior~\cite{ds_slam,dynaslam,dynamicslam,dynaslam2,vdoslam} or optical flow motion~\cite{sun2018motion,cheng2019improving,flowfusion} to remove the dynamic objects. 
However, all these methods often exhibit a domain gap when applied in real-world environments, leading to the introduction of prediction errors. 
In this paper, we propose a motion mask generation strategy that complements the semantic segmentation mask with warping optical flow masks~\cite{raft,gma}, which is beneficial for reconstructing more accurate static scene maps and reducing observation error.
\subsection{RGB-D SLAM with neural implicit representation}
Neural implicit scene representations, also known as neural fields~\cite{mildenhall2021nerf}, have garnered significant interest in RGB-D SLAM due to their expressive capacity and minimal memory requirements. iMap~\cite{imap} firstly adopts a single MLP representation to jointly optimize camera pose and implicit map throughout the tracking and mapping stages. However, it suffers from representation forgetting problems and fails to produce detailed scene geometry. DI-Fusion~\cite{difusion} encodes the scene prior in a latent space and optimizes a feature grid, but it leads to poor reconstruction quality replete with holes. NICE-SLAM~\cite{nice_slam} leverages a multi-level feature grid enhancing scene representation fidelity and utilizes a local feature update strategy to reduce network forgetting. However, it remains memory-intensive and lacks real-time capability. More recently, existing methods like Vox-Fusion~\cite{voxfusion}, Co-SLAM~\cite{co-slam}, and ESLAM~\cite{eslam} explore sparse encoding or tri-plane representation strategy to improve the quality of scene reconstruction and the system's execution efficiency. 
All these methods have demonstrated impressive results based on the strong assumptions of static scene conditions. The robustness of these systems significantly decreases when dynamic objects appear in the environment. Our SLAM system aims to enhance the accuracy and robustness of pose estimation under dynamic environments, which can expand the application range for the NeRF-based RGB-D SLAM system.
\subsection{Dynamic objects decomposition in NeRFs}
As the field of NeRF continues to advance, some researchers are attempting to address the problem of novel view synthesis in the presence of dynamic objects. One kind of solution is to decompose the static background and dynamic objects with different neural radiance fields like~\cite{nerf_in_the_wild, nerfies, dnerf, hypernerf, DynNeRF, chen2022flow,d2nerf}. The time dimension will be encoded in latent space, and novel view synthesis is conducted in canonical space. Although these space-time synthesis results are impressive, these techniques rely on precise camera pose input. Robust-Dynrf~\cite{rodyraf} jointly estimate the static and dynamic radiance fields along with the camera parameters (poses and focal length), which can achieve the unknown camera pose training. However, it can not directly apply to RGB-D SLAM system for large-scale tracking and mapping. Another kind of solution is to ignore the dynamic objects' influence by utilizing robust loss and optical flow like~\cite{chen2022flow,robustnerf}. Compared to the dynamic NeRF problem, we often focus on the accuracy of pose estimation and the quality of static reconstruction without a long training period. Thus, we also ignore modeling dynamic objects and propose a robust loss function with a novel optimization strategy to recover the static scene map.

\begin{figure*}[t]
\scriptsize
\vspace{-2mm} 
\centering
\includegraphics[width=0.85\textwidth]{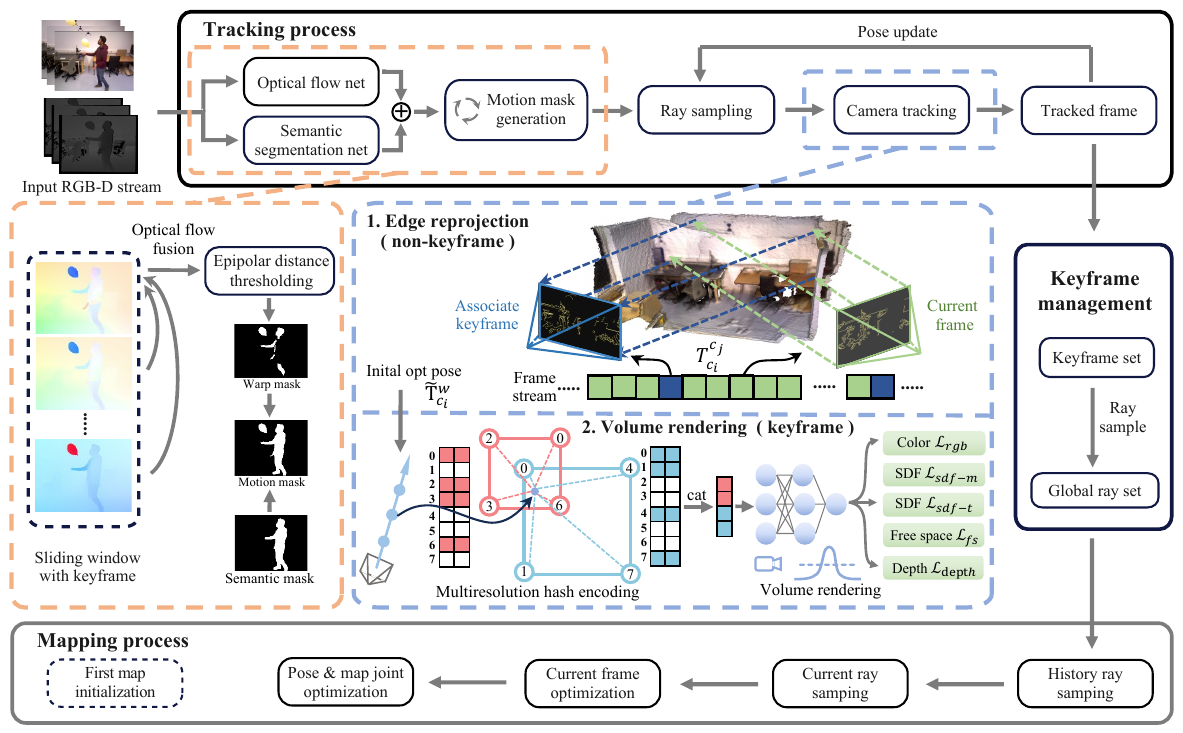} 
\vspace{-4mm} 
\caption{\textbf{The schematic illustration of the proposed method.} Given a series of RGB-D frames, we simultaneously construct the implicit map and camera pose via multi-resolution hash gird with the geometric loss $\mathcal{L}_{sdf\text{-}m}, \mathcal{L}_{sdf\text{-}t}, \mathcal{L}_{fs}, \mathcal{L}_{depth}$, color loss $\mathcal{L}_{color}$, and edge warp loss $\mathcal{L}_{edge}$. 
}
\label{fg:pipeline}
\vspace{-0.1in}
\end{figure*}

\section{Method}
\label{sec:approach}
Given a sequence of RGB-D frames $\{I_i, D_i\}_{i=1}^N, I_i \in \mathbb{R}^3, D_i \in \mathbb{R}$, our method (Fig.~\ref{fg:pipeline}) aims to simultaneously recover camera poses $\{\xi_i\}_{i=1}^N, \xi_t \in \mathbb{SE}(3)$ and reconstruct the static 3D scene map represented by neural radiance fields in dynamic environments. 
Similar to most modern SLAM systems~\cite{PTAM2007, DTAM_Newcombe2011}, our system comprises two distinct processes: the tracking process as the frontend and the mapping process as the backend, combined with keyframe management $\{F_k\}_{k=1}^M$ and neural implicit map $f_\theta$. 
Invalid sampling rays within dynamic objects are filtered out using a motion mask generation approach. 
Contrary to the conventional constant-speed motion model in most systems, we introduce an edge warp loss for optimization in non-keyframes to enhance the robustness of pose estimation. 
Furthermore, keyframe poses and the implicit map representations are iteratively optimized using differentiable rendering.
\subsection{Implicit map representation}
\label{sec:nerf_representation}
We introduce two components of our implicit map representation: an efficient multi-resolution hash encoding $\mathbb{V}_{\alpha}$ to encode the geometric information of the scene, and individual tiny MLP decoders $f_{\phi}$ to render the color and depth information with truncated signed distance (TSDF) prediction.

\paragraph{Multi-resolution hash encoding}
We use a multi resolution hash-based feature grid $\mathbb{V}_{\alpha} = \{V^l_\alpha\}_{l=1}^L$ and individual shallow MLPs to represent the implicit map following Instant-NGP~\cite{instantngp}. The spatial resolution of each level is progressively set between the coarsest resolution, denoted as $R_{min}$, and the finest resolution, represented as $R_{max}$. Given a sampled point $\mathbf{x}$ in 3D space, we compute the interpolate feature $V^l_\alpha(\mathbf{x})$ from each level via trilinear interpolation. To obtain more complementary geometric information, we concat the encoding features from all levels as the input of the MLPs decoder.
While simple MLPs can lead to the issue of catastrophic forgetting~\cite{imap,nice_slam}, 
this \textbf{mechanism of forgetfulness} can be leveraged to eliminate historical dynamic objects.

\paragraph{Color and depth rendering}
To obtain the final formulation of implicit map representation, we adopt a two-layer shallow MLP to predict the geometric and appearance information, respectively. The geometry decoder outputs the predicted SDF value $s$ and a feature vector $\mathbf{h}$ at the point $\mathbf{x}$. The appearance decoder outputs the predicted RGB value $c$. Similar to Co-SLAM~\cite{co-slam}, we joint encode the coordinate encoding $\gamma(\mathbf{x})$ and parametric encoding $V_{\alpha}$ as:
\begin{equation}
f_\beta\left(\gamma(\mathbf{x}), V_\alpha(\mathbf{x})\right) \mapsto(\mathbf{h}, s) , \quad f_\phi\left(\gamma(\mathbf{x}), \mathbf{h}\right) \mapsto \mathbf{c},
\end{equation}
where $\{\alpha, \beta, \phi\}$ are the learnable parameters. Following the volume rendering method in NeRF~\cite{nerf}, we accumulate the predicted values along the viewing ray $\mathbf{r}$ at the current estimation pose $\xi_i$ to render the color and depth value as:
\begin{equation}
\resizebox{0.47\textwidth}{!}{$
\hat{C}(\mathbf{r})=\frac{1}{\sum_{i=1}^M w_i} \sum_{i=1}^M w_i \mathbf{c}_i, \quad \hat{D}(\mathbf{r})=\frac{1}{\sum_{i=1}^M w_i} \sum_{i=1}^M w_i z_i,
$}
\end{equation}
where $w_i$ is the computed weight along the ray, $\mathbf{c}_i$ and $z_i$ are the color and depth value of the sampling point $\mathbf{x}_i$. Since we do not directly predict voxel density $\sigma$ like NeRF, here we need to convert the SDF values $s_i$ into weights $w_i$. Thus, we employ a straightforward bell-shaped function~\cite{neuralrgbd}, formulated as the product of two sigmoid functions $\sigma(\cdot)$.
\begin{equation}
\resizebox{0.47\textwidth}{!}{$
w_i=\sigma\left(\frac{s_i}{t r}\right) \sigma\left(-\frac{s_i}{t r}\right), \quad \hat{D}_{var}(\mathbf{r})=\frac{1}{\sum_{i=1}^M w_i} \sum_{i=1}^M w_i (\hat{D} - z_i)^2,
$}
\end{equation}
where $tr$ denotes the truncation distance with TSDF prediction, $\hat{D}_{var}$ is the depth variance along this ray. When possessing GT depth values, we opt for uniform point sampling near the surface rather than employing importance sampling, with the aim of enhancing the efficiency of point sampling.

\subsection{Motion mask generation}
\label{sec:motion_mask}
For each input keyframe, we select its associated keyframes within a sliding window to compute the dense optical flow warping set $\mathcal{S}$. 
Note that optical flow estimation is conducted solely on keyframes, thereby optimizing system efficiency. To separate the ego-motion from dynamic objects, we additionally estimate the fundamental matrix $\mF$ with inliers sampled from the matching set $\mathcal{S}$. 
Given any matching points $\vo_{ji},\vo_{ki}$ within $\mathcal{S}$, we utilize matrix $\mF$ to compute the Sampson distance between corresponding points and their epipolar lines. 
By setting a suitable threshold $e_{th}$, we derive the warp mask $\widehat{\mathcal{M}}_{j,k}^{wf}$ corresponding to dynamic objects as:
\begin{equation}
\resizebox{0.47\textwidth}{!}{$
\widehat{\mathcal{M}}_{j,k}^{wf} : \left\{\bigcap_{i=1}^M\displaystyle \1(\frac{\vo_{ji}^{T} \mF \vo_{ki}}{\sqrt{A^2+B^2}} < e_{th}) \otimes \displaystyle \mI_{m\times n} \, \bigg| \, \forall \left(\vo_{ji}, \vo_{ki}\right) \in \mathcal{S}\right\}$}
\end{equation}
where $A, B$ denotes the coefficients of the epipolar line, and $m, n$ represents the size of the warp mask, aligning with the current frame image's dimensions. Additionally, $j$ and $k$ stand for the keyframe ID, illustrating the optical flow mask warping process from the k-th to the j-th keyframe. As illustrated in Fig.~\ref{fg:pipeline}, to derive a more precise motion mask, we consider the spatial coherence of dynamic object motions within a sliding window of length $N$ and iteratively optimize the current motion mask. Subsequently, we integrate the warp mask and segment mask to derive the final motion mask $\widehat{\mathcal{M}}_j$ as:
\begin{equation}
\resizebox{0.45\textwidth}{!}{$
\widehat{\mathcal{M}}_j=\widehat{\mathcal{M}}_{j,k}^{wf} \otimes \widehat{\mathcal{M}}_{j,k-1}^{wf} \otimes \widehat{\mathcal{M}}_{j,k-2}^{wf} \cdots \otimes\widehat{\mathcal{M}}_{j,k-N}^{wf} \cup \widehat{\mathcal{M}}_{j}^{sg},$}
\end{equation}
where $\otimes$ represents the mask fusion operation, which is applied when pixels corresponding to a specific motion mask have been continuously observed for a duration exceeding a certain threshold $o_{th}$ within a sliding window. Note that we do not focus on the specific structure of the segment or optical flow network. Instead, we aim to introduce a general motion mask fusing method for application in NeRF-based SLAMs. We believe that there is potential for integrating this approach into any visual SLAM system.

\subsection{Joint optimization}
\label{sec:joint_optimization}
We introduce the details on optimizing the implicit scene representation and camera pose. Given a set of frames $\mathcal{F}$, we only predict the current camera pose represented with lie algebra $\xi_i$ in tracking process. Moreover, we utilize the global bundle adjustment (GBA)~\cite{nerfmm,lin2021barf,bian2023nope} to jointly optimize the sampled camera pose and the implicit mapping.

\subsubsection{Photometric rendering loss} 
To jointly optimize the scene representation and camera pose, we render depth and color in independent view as Eq.~\ref{eq:render loss} comparing with the proposed ground truth map: 
\begin{equation}
\label{eq:render loss}
\begin{aligned}
    \mathcal{L}_{rgb} &= \frac{1}{M}\sum_{i=1}^M {\Big\| \left(\hat{C}(\mathbf{r}) - C(\mathbf{r}) \right) \cdot  \widehat{\mathcal{M}}_i(\mathbf{r}) \Big\|}^2_2, \\
    \mathcal{L}_{depth} &= \frac{1}{N_d}\sum_{\mathbf{r} \in N_d} {\bigg\| \left(\frac{\hat{D}(\mathbf{r}) - D(\mathbf{r})}{\sqrt{\hat{D}_{var}(\mathbf{r})}} \right) \cdot \widehat{\mathcal{M}}_i(\mathbf{r}) \bigg\|^2_2},
\end{aligned}
\end{equation}
where $C(\mathbf{r})$ and $D(\mathbf{r})$ denote the ground truth color and depth map corresponding with the given pose,s respectively. $M$ represents the number of sampled pixels in the current image. Note that only rays with valid depth value $N_d$ are considered in $\mathcal{L}_{depth}$. In contrast to existing methods, we introduce the motion mask $\widehat{\mathcal{M}}_j$ to remove sampled pixels within the dynamic object region effectively. Moreover, to improve the robustness of pose estimation, we add the depth variance $\hat{D}_{var}$ to reduce the weight of depth outliers.

\subsubsection{Geometric constraints} Following the practice~\cite{neuralrgbd}, assuming a batch of rays $M$ within valid motion mask regions are sampled, we directly leverage the free space loss with truncation $tr$ to restrict the SDF values $s(\mathbf{x_i})$ as:
\begin{equation}
\label{eq:fs loss}
    \mathcal{L}_{fs} = \frac{1}{M}\sum_{i=1}^M \frac{1}{|\mathcal{R}_{fs}|} \sum_{i \in \mathcal{R}_{fs}} (s(\mathbf{x}_i)-tr)^2, \quad [u_i, v_i] \subseteq (\widehat{\mathcal{M}}_i = 1).
\end{equation}
It is unreasonable to employ a fixed truncation value to optimize camera pose and SDF values in dynamic environments simultaneously. To reduce the artifacts in occluded areas and enhance the accuracy of reconstruction, we further divide the entire truncation region near the surface into middle and tail truncation regions inspired by ESLAM~\cite{eslam} as:
\begin{equation}
\label{eq:sdf loss}
    \mathcal{L}_{sdf} = \frac{1}{M}\sum_{i=1}^M \frac{1}{|\mathcal{R}_{tr}|} \sum_{i \in \mathcal{R}_{tr}} \left(s(\mathbf{x}_i) - (D[u_i,v_i]-T\cdot {tr})\right)^2,
\end{equation}
where $T$ denotes the ratio of the entire truncation length occupied by the middle truncation, $[u_i, v_i] \subseteq (\widehat{\mathcal{M}}_i = 1)$. Note that we use the different weights to adjust the importance of middle and tail truncation in camera tracking and mapping process.
The overall loss function is finally formulated as the following minimization, 
\begin{equation}
\label{eq:keyframe_opt}
    \mathcal{P}^* = \underset{\mathcal{P}}{\mathop{\arg\min}} \, \, \lambda_1\mathcal{L}_{rgb} + \lambda_2\mathcal{L}_{depth} + \lambda_3\mathcal{L}_{fs}
    + \lambda_4\mathcal{L}_{sdf\text{-}m} + \lambda_5\mathcal{L}_{sdf\text{-}t},
\end{equation}
where $\mathcal{P} = \{\theta, \phi, \alpha, \beta, \gamma, \mathbf{\xi}_i\}$ is the list of parameters being optimized, including fields feature, decoders, and camera pose. 
\subsubsection{Camera tracking process}
The construction of implicit maps within dynamic scenes often encounters substantial noise and frequently exhibits a lack of global consistency. Existing methods~\cite{nice_slam,co-slam,eslam,dimslam} rely solely on rendering loss for camera pose optimization, which makes the system vulnerable and prone to tracking failures. To solve this problem, we introduce edge warp loss to enhance geometry consistency in data association between adjacent frames.

\noindent \textbf{Edge reprojection loss.} 
For a 2D pixel $p$ in frame $i$, we first define the warp operation in a similar spirit as DIM-SLAM~\cite{dimslam} to reproject it onto frame $j$ as follows:
\begin{equation}
\vp_{i\rightarrow j}=f_{warp}\left(\xi_{j i}, \vp_i, D(\vp_i)\right)=\mK \mT_{j i} \left(\mK^{-1}D(\vp_i)\vp_i^{homo}\right),
\end{equation}
where $\mK$ and $\mT_{ji}$ represent the intrinsic matrix and the transformation matrix between frame $i$ and frame $j$, respectively. $\vp_i^{homo} = (u,v,1)$ is the homogeneous coordinate of $\vp_{i}$. Since the edge are detected once and do not change forwards, we can precompute the distance map (DT)~\cite{dtmap} to describe the projection error with the closest edge. For a edge set $\mathcal{E}_i$ in frame $i$, we define the edge loss $\mathcal{L}_{edge}$ as 
\begin{equation}
    \mathcal{L}_{edge} = \sum_{\vp_i \in \mathcal{E}_i} \rho(\mathcal{D}_j(f_{warp}\left(\xi_{j i}, \vp_i, D(\vp_i)\right))\cdot \widehat{\mathcal{M}}_j),
\end{equation}
where $\mathcal{D}_j$ denotes the DT map in frame $j$, and the $\rho$ is a Huber weight function to reduce the influence of large residuals. Moreover, we drop a potential outlier if the projection distance error is greater than $\delta_e$. The pose optimization problem is finally formulated as the following minimization,
\begin{equation}
\label{eq:nonkey_opt}
    \xi_{j i}^*=\underset{\xi_{j i}}{\operatorname{argmin}} \, \, \lambda\mathcal{L}_{edge} , \quad  \text{if} \, \, j \notin \mathcal{K}
\end{equation}
To further improve the accuracy and stability of pose estimation, we employ distinct methods for tracking keyframes and non-keyframes in dynamic scenes. 
Keyframe pose estimation utilizes the edge loss to establish the initial pose, followed by optimization (Eq.~\ref{eq:keyframe_opt}). 
For non-keyframe pose estimation, we optimize the current frame's pose related to the nearest keyframe (Eq.~\ref{eq:nonkey_opt}).

\section{Experiments}
\label{sec:exp}

\noindent \textbf{Datasets.} 
We evaluate our method on two real-world public datasets: \textit{TUM RGB-D} dataset~\cite{tumbenchmark} and \textit{BONN RGB-D Dynamic} dataset~\cite{refusion}. Both datasets capture indoor scenes using a handheld camera and provide the ground-truth trajectory. 

\noindent \textbf{Metrics.} 
\label{sec:exp_metrics}
For evaluating pose estimation, we adopt the RMSE and STD of Absolute Trajectory Error (ATE)~\cite{tumbenchmark}. The estimated trajectory is oriented to align with the ground truth trajectory using the unit quaternions algorithm~\cite{horn1987closed} before evaluation. 
We also use three metrics which are widely used for scene reconstruction evaluation following~\cite{nice_slam,dimslam}: (i) \textit{Accuracy} (cm), (ii) \textit{Completion} (cm), (iii) \textit{Completion Ratio} ($<$ 5cm \%). Since the BONN-RGBD only provided the ground truth point cloud, we randomly sampled the 200,000 points from both the ground truth point cloud and the reconstructed mesh surface to compute the metrics. We remove unobserved regions that are outside of any camera's viewing frustum and conduct extra mesh culling to remove the noisy points external to the target scene~\cite{co-slam}.

\noindent \textbf{Implementation details.} We adopt Co-SLAM \cite{co-slam} as the baseline in our experiments and run our RoDyn-SLAM on an high-performance workstation with a 3.4GHz Intel Core i7-13700K CPU and RTX 3090Ti GPU at 10 FPS (without optical flow mask) on the Tum datasets, which takes roughly 4GB of memory in total. Specific to implementation details, we sample $N_t$ = 1024 rays and $N_p$ = 85 points along each camera ray with 20 iterations for tracking and 2048 pixels from every 5 $th$ frames for global bundle adjustment. 
We set loss weight $\lambda_1$ = 1.0, $\lambda_2$ = 0.1, $\lambda_3$ = 10, $\lambda_4$ = 2000, $\lambda_5$ = 500 to train our model with Adam~\cite{kingma2014adam} optimizer. In the motion mask generation method, we utilize Oneformer~\cite{oneformer} for semantic segmentation prior generation and RAFT-GMA~\cite{gma} for optical flow prediction. In the edge extraction process, we utilize the Canny~\cite{canny1986computational} edge detection algorithm with double-threshold. For the sake of comparison fairness, we employ the same keyframe insertion strategy as Co-SLAM~\cite{co-slam}.

\subsection{Evaluation of generating motion mask}
\begin{figure*}[t]
\scriptsize
\vspace{-3mm} 
\centering
\includegraphics[width=0.96\textwidth]{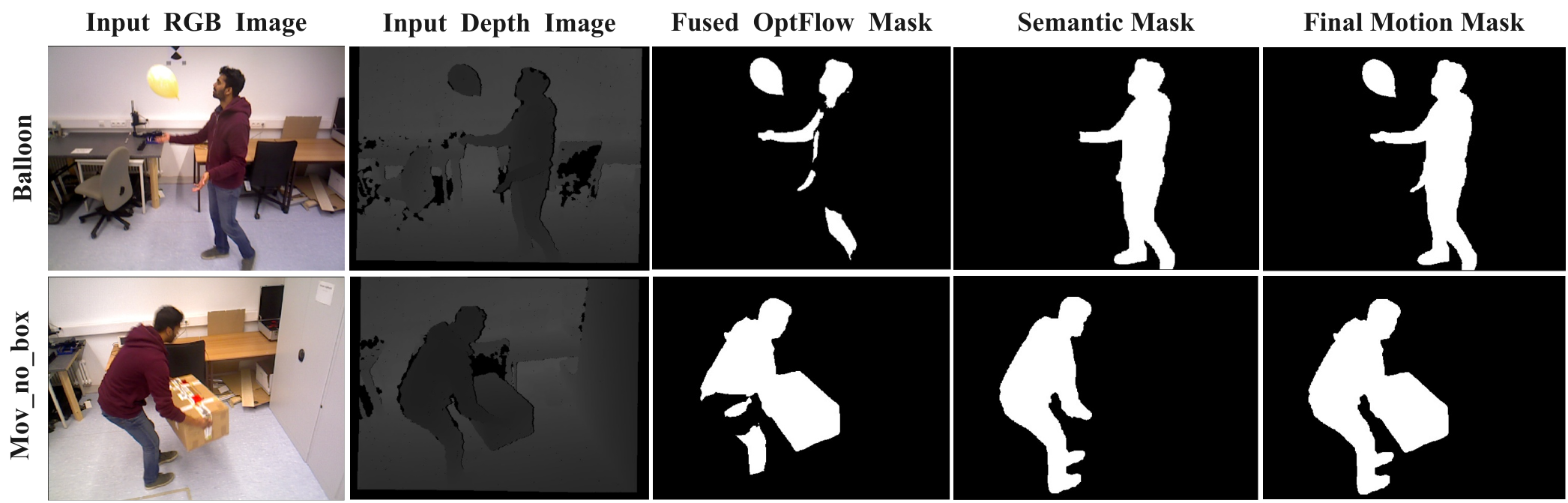} 
\vspace{-4mm}
\caption{\textbf{Qualitative results of the generation motion mask.} By iteratively optimizing the optical flow mask, the fused optical mask can be more precise without noises. The semantic mask can only identify dynamic objects within predefined categories. The best results are obtained with our method.}
\label{fg:motion_mask}
\vspace{-0.1in}
\end{figure*}
\begin{table*}[t]
\vspace{-3mm}
\caption{Quantitative results on several dynamic scene sequences in the \textit{BONN-RGBD} dataset. ``X" denotes the tracking failures. The best results are \textbf{bolded}, and the second best results are indicated with an underline.}
\vspace{-4mm}
\begin{center}
\begin{tabular}{clcccccc}
\toprule
& & \texttt{ball}& \texttt{ball2}& \texttt{ps\_trk} & \texttt{ps\_trk2}& \texttt{mv\_box2} & Avg.\\
\midrule
\multirow{3}{*}{ReFusion~\cite{refusion}} & \textbf{Acc.}[cm]$\downarrow$ &\textbf{8.20}&\textbf{7.85}&46.89&78.47&\textbf{9.07}&30.10\\
                      & \textbf{Comp.}[cm]$\downarrow$& 12.58 & \underline{11.69}& \underline{104.04}& 166.63&13.09&61.61 \\
                      & \textbf{Comp. Ratio}[$\leq5$cm$\%$]$\uparrow$&31.57&32.18&13.93&10.55&35.51&24.75\\
\midrule
\multirow{3}{*}{iMAP*~\cite{imap}} & \textbf{Acc.}[cm]$\downarrow$ &16.68&31.20&35.38&54.16&17.01&30.89\\
                      & \textbf{Comp.}[cm]$\downarrow$& 27.32 & 30.14& 201.38& \underline{107.28}&20.499& 77.32 \\
                      & \textbf{Comp. Ratio}[$\leq5$cm$\%$]$\uparrow$&25.68&21.91&11.54&12.63&24.86&19.32\\
\midrule
\multirow{3}{*}{NICE-SLAM~\cite{nice_slam}} & \textbf{Acc.}[cm]$\downarrow$& X &24.30&43.11&74.92& 17.56 & 39.97\\
                      & \textbf{Comp.}[cm]$\downarrow$ & X & 16.65 & 117.95 & 172.20 & 18.19  & 81.25  \\
                      & \textbf{Comp. Ratio}[$\leq5$cm$\%$]$\uparrow$& X &29.68&15.89&13.96& 32.18 & 22.93 \\
\midrule
\multirow{3}{*}{Vox-Fusion~\cite{voxfusion}} & \textbf{Acc.}[cm]$\downarrow$&85.70&89.27&208.03&162.61&40.64&117.25\\
                      & \textbf{Comp.}[cm]$\downarrow$&55.01&29.78&279.42&229.79&28.40&124.48\\
                      & \textbf{Comp. Ratio}[$\leq5$cm$\%$]$\uparrow$&3.88&11.76&2.17&4.55&14.69&7.41\\
\midrule
\multirow{3}{*}{Co-SLAM~\cite{co-slam}} & \textbf{Acc.}[cm]$\downarrow$&10.61&14.49&\underline{26.46}&\underline{26.00}&12.73&\underline{18.06}\\
                      & \textbf{Comp.}[cm]$\downarrow$&10.65&40.23&124.86&118.35&10.22&\underline{60.86}\\
                      & \textbf{Comp. Ratio}[$\leq5$cm$\%$]$\uparrow$&34.10&3.21&2.05&2.90&39.10&16.27\\
                      
\midrule
\multirow{3}{*}{ESLAM~\cite{eslam}} & \textbf{Acc.}[cm]$\downarrow$&17.17&26.82&59.18&89.22&12.32&40.94\\
& \textbf{Comp.}[cm]$\downarrow$&\underline{9.11}&13.58&145.78&186.65&\underline{10.03}&73.03\\
& \textbf{Comp. Ratio}[$\leq5$cm$\%$]$\uparrow$&\underline{47.44}&\textbf{47.94}&\underline{20.53}&\underline{17.33}&\underline{41.41}&\underline{34.93}\\
\midrule
\multirow{3}{*}{Ours(RoDyn-SLAM)} & \textbf{Acc.}[cm]$\downarrow$&\underline{10.60}&\underline{13.36}&\textbf{10.21}&\textbf{13.77}&\underline{11.34}&\textbf{11.86}\\
                      & \textbf{Comp.}[cm]$\downarrow$&\textbf{7.15}&\textbf{7.87}&\textbf{27.70}&\textbf{18.97}&\textbf{6.86}&\textbf{13.71}\\
                      & \textbf{Comp. Ratio}[$\leq5$cm$\%$]$\uparrow$&\textbf{47.58}&\underline{40.91}&\textbf{34.13}&\textbf{32.59}&\textbf{45.37}&\textbf{40.12}\\
\bottomrule
\end{tabular}
\label{tb:recon_bonn}
\end{center}
\end{table*}

Fig.~\ref{fg:motion_mask} shows the qualitative results of the generated motion mask. We evaluated our method on the \texttt{balloon} and \texttt{move\_no\_box2} sequence of the \textit{BONN} dataset. In these sequences, in addition to the movement of the person, there are also other dynamic objects associated with the person, such as balloons and boxes.
As shown in Fig.~\ref{fg:motion_mask} final mask part, our methods can significantly improve the accuracy of motion mask segmentation and effectively mitigate both false positives and false negatives issues in motion segmentation. 

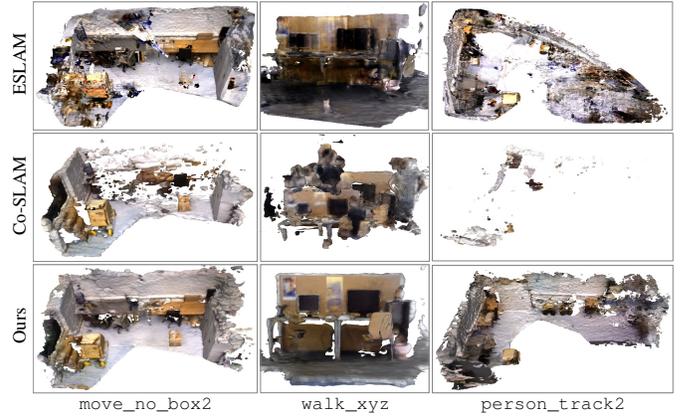
\begin{figure}[t]
\raggedleft
\includegraphics[width=0.47\textwidth]{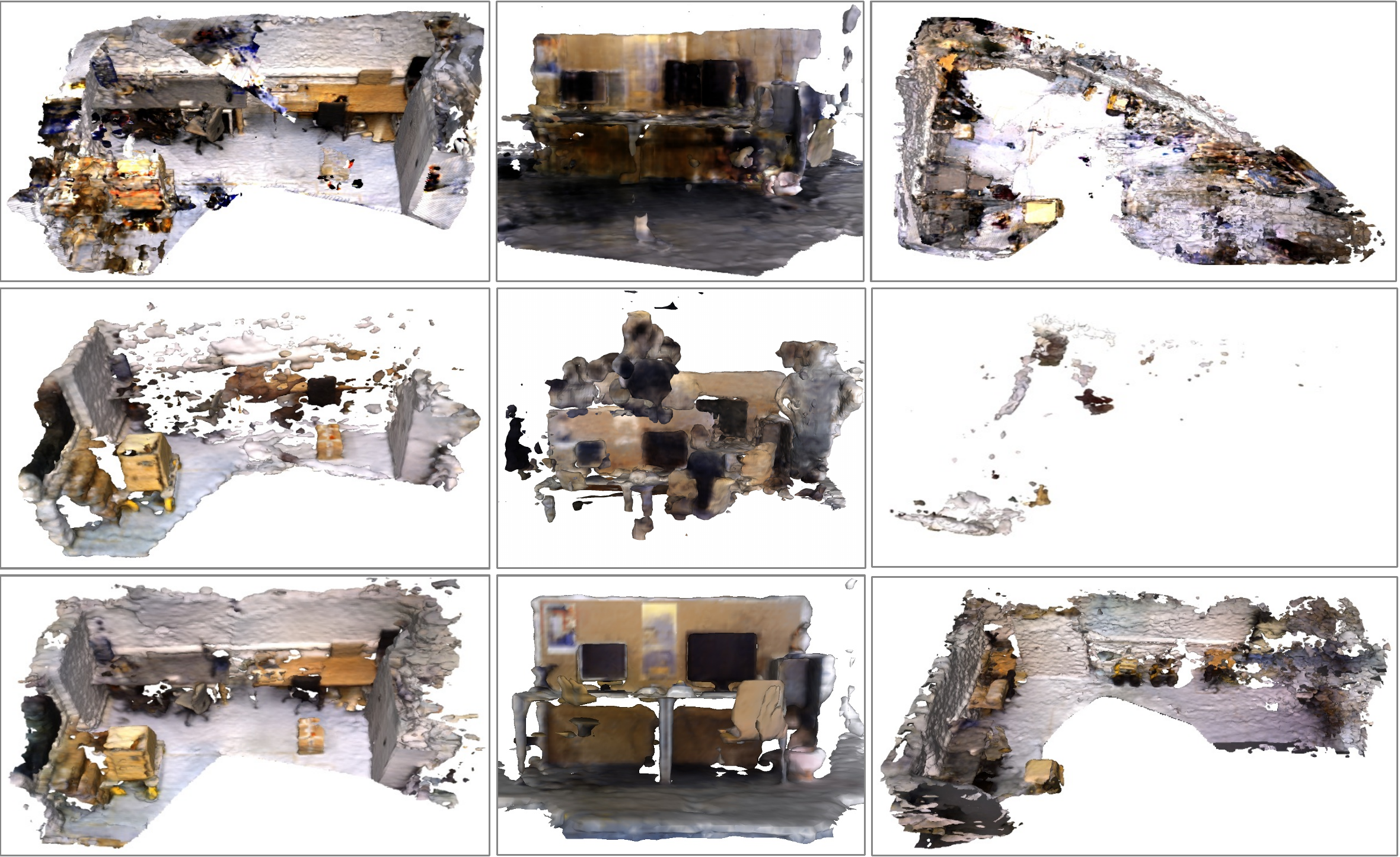}
\put(-225,-6){\scriptsize{\texttt{move\_no\_box2}}}
\put(-141,-6){\scriptsize{\texttt{walk\_xyz}}}
\put(-73,-6){\scriptsize{\texttt{person\_track2} }}
\put(-250,113){\scriptsize{\rotatebox{90}{ESLAM}}}
\put(-250,60){\scriptsize{\rotatebox{90}{Co-SLAM}}}
\put(-250,19){\scriptsize{\rotatebox{90}{Ours}}}
\vspace{-1mm}
 \caption{\textbf{Visual comparison of the reconstructed meshes on the \textit{BONN} and \textit{TUM RGB-D} datasets.} Our results are more complete and accurate without the dynamic object floaters.}
\label{fig:vis_compare}
\vspace{-1mm}
\end{figure}

\subsection{Evaluation of mapping and tracking}
\vspace{-0.05in}
\paragraph{Mapping} 
To better demonstrate the performance of our proposed system in dynamic scenes, we evaluate the mapping results from both qualitative and quantitative perspectives. Since the majority of dynamic scene datasets do not provide ground truth for static scene reconstruction, we adopt the \textit{BONN} dataset to conduct quantitative analysis experiments. We compare our RoDyn-SLAM method against traditional dynamic SLAM method like ReFusion~\cite{refusion} and current state-of-the-art NeRF-based methods with RGB-D sensors, including NICE-SLAM~\cite{nice_slam}, iMap~\cite{imap}, Vox-Fusion~\cite{voxfusion}, ESLAM~\cite{eslam}, and Co-SLAM~\cite{co-slam}, which are open source. The evaluation metrics have been mentioned above at the beginning of Section~\ref{sec:exp}.

\setlength{\tabcolsep}{3.5pt}
\begin{table*}[!t]
\vspace{-2mm}
\caption{Camera tracking results on several dynamic and static scene sequences in the \textit{TUM RGB-D} dataset. 
``$*$" denotes the version reproduced by NICE-SLAM. ``-" denote the absence of mention.
The metric unit is [cm]. }
\vspace{-4mm}
\begin{center}
\footnotesize
\begin{tabular}{l|c|cc|cc|cc|cc|cc|cc|cc}
\toprule
\multirow{2}{*}{Methods} &\multirow{2}{*}{Dense} &\multicolumn{8}{c|}{Dynamic} &\multicolumn{4}{c|}{Static} &\multicolumn{2}{c}{\multirow{2}{*}{Avg.}}\\
&  & \multicolumn{2}{c}{\texttt{f3/wk\_xyz}} & \multicolumn{2}{c}{\texttt{f3/wk\_hf}} & \multicolumn{2}{c}{\texttt{f3/wk\_st}} & \multicolumn{2}{c|}{\texttt{f3/st\_hf}}  & \multicolumn{2}{c}{\texttt{f1/xyz}} & \multicolumn{2}{c|}{\texttt{f1/rpy}} \\
\midrule
\textit{Traditional SLAM methods}  & \textit{T/F} & \textit{ATE} &\textit{S.D.}  & \textit{ATE} &\textit{S.D.}   & \textit{ATE} &\textit{S.D.}  &  \textit{ATE} &\textit{S.D.}   & \textit{ATE} &\textit{S.D.}  & \textit{ATE} &\textit{S.D.} & \textit{ATE} &\textit{S.D.}\\
ORB-SLAM3\cite{orbslam3} &\xmark & 28.1 & 12.2 & 30.5 & 9.0 &2.0 & 1.1 & \textbf{2.6}  & 1.6 &\textbf{1.1}  &0.6 &2.2  &1.3 &11.1  &4.3\\
DVO-SLAM\cite{dvoslam} &\cmark & 59.7 & - & 52.9 & - &21.2 & - & 6.2  & - &\textbf{1.1}  &- &\textbf{2.0}  &- &22.9  &-\\
DynaSLAM\cite{dynaslam} &\xmark & \textbf{1.7} & -  & \textbf{2.6} & - & \textbf{0.7} & -  & 2.8 & - & -  &- &-  &- &\textbf{2.0}  &-\\
ReFusion\cite{refusion} &\cmark & 9.9 & -  & 10.4 & -  &1.7 & -  & 11.0 & - &- &- &-  &- &8.3  &-\\
\midrule
\textit{NeRF based SLAM methods} &\textit{T/F} &\textit{ATE} &\textit{S.D.}  & \textit{ATE} &\textit{S.D.}   & \textit{ATE} &\textit{S.D.}  &  \textit{ATE} &\textit{S.D.}   & \textit{ATE} &\textit{S.D.}  & \textit{ATE} &\textit{S.D.} & \textit{ATE} &\textit{S.D.}\\
iMAP*\cite{imap} &\cmark &111.5  & 43.9  & X  &X  &137.3  & 21.7 & 93.0  & 35.3 & 7.9  &7.3 &16.0  & 13.8 &73.2 & 24.4\\
NICE-SLAM\cite{nice_slam} &\cmark &113.8  & 42.9 & X & X  &88.2  & 27.8 & 45.0 &14.4 & 4.6 & 3.8 &3.4  & 2.5 & 51 & 18.3\\
Vox-Fusion\cite{voxfusion} &\cmark &146.6   & 32.1  & X & X &109.9  & 25.5 & 89.1 & 28.5 &1.8  & 0.9 &4.3   & 3.0 &70.4 &18\\
Co-SLAM\cite{co-slam} &\cmark &51.8 & 25.3  & 105.1 & 42.0 &49.5 & 10.8  & 4.7 & 2.2 & 2.3 & 1.2 &3.9 & 2.8 &36.3 &14.1\\
ESLAM\cite{eslam} &\cmark &45.7 & 28.5 & 60.8 & 27.9 &93.6 & 20.7  & \textbf{3.6} &\textbf{1.6} & \textbf{1.1} & \textbf{0.6} &\textbf{2.2} & \textbf{1.2} &34.5 &13.5\\
RoDyn-SLAM(Ours) &\cmark & \textbf{8.3} & \textbf{5.5} & \textbf{5.6} & \textbf{2.8} & \textbf{1.7}  & \textbf{0.9} & 4.4 & 2.2 & 1.5 & 0.8 & 2.8 & 1.5 &\textbf{4.1} &\textbf{2.3}\\
\bottomrule 
\end{tabular}
\label{table:tum}
\end{center}
\end{table*}

As shown in Tab.~\ref{tb:recon_bonn}, our method outperforms most of the neural RGB-D slam systems on accuracy and completion. To improve the accuracy of pose estimation, we filter the invalid depth, which may reduce the accuracy metric on mapping evaluation.
The visual comparison of reconstructed meshes with other methods~\cite{eslam, co-slam} is provided in Fig.~\ref{fig:vis_compare}. Note that the TUM dataset does not provide ground truth meshes for evaluating mapping quality. Our methods can generate a more accurate static mesh than other compared methods. Since the baseline methods~\cite{co-slam} adopt the hash encoding to represent the implicit map, it may exacerbate the issue of the hash collisions in dynamic scenes and generate the hole in the reconstruction map. 

\paragraph{Tracking}
To evaluate the accuracy of camera tracking in dynamic scenes, we compare our methods with the recent neural RGB-D SLAM methods and traditional SLAM methods like ORB-SLAM3~\cite{orbslam3}, DVO-SLAM~\cite{dvoslam}, Droid-SLAM~\cite{teed2021droid},
and traditional dynamic SLAM like DynaSLAM~\cite{dynaslam}, 
and ReFusion~\cite{refusion}. 

As shown in Tab.~\ref{table:tum}, we report the results on three highly dynamic sequences, one slightly dynamic sequence, and two static sequences from TUM RGB-D dataset. Our system achieves advanced tracking performance owing to the motion mask filter and edge-based optimization algorithm under dynamic environment. Compared with our baseline methods Co-SLAM~\cite{co-slam}, our method does not compromise the performance of the original SLAM methods in terms of tracking and mapping in static scenes. In fact, it achieves competitive results. Notably, our proposed optimization algorithm is not restricted to a specific slam system. Thus, it can also be applied to other neural rgb-d slam methods to improve the data association between the inter-frame. We have also evaluated the tracking performance on the more complex and challenging BONN RGB-D dataset, as illustrated in Tab.~\ref{table:bonn}. 
In more complex and challenging scenarios, our method has achieved superior results. While there is still some gap compared to the more mature and robust traditional dynamic SLAM methods, our systems can drive the dense and textural reconstruction map to finish the more complex robotic navigation tasks. 

\setlength{\tabcolsep}{3.5pt}
\begin{table*}[!t]
\vspace{-2mm}
\caption{Camera tracking results on several dynamic scene sequences in the \textit{BONN RGB-D} dataset. 
``$*$" denotes the version reproduced by NICE-SLAM. ``-" denote the absence of mention, respectively. The metric unit is [cm].}
\vspace{-4mm}
\begin{center}
\begin{tabular}{l|c|cc|cc|cc|cc|cc|cc|cc}
\toprule
\multicolumn{1}{l}{Methods} & \multicolumn{1}{c}{Dense} & \multicolumn{2}{c}{\texttt{balloon}} & \multicolumn{2}{c}{\texttt{balloon2}} & \multicolumn{2}{c}{\texttt{ps\_track}} & \multicolumn{2}{c}{\texttt{ps\_track2}} & \multicolumn{2}{c}{\texttt{ball\_track}} & \multicolumn{2}{c}{\texttt{mv\_box2}}  & \multicolumn{2}{c}{Avg.} \\
\midrule
\textit{Traditional SLAM methods} & \textit{T/F}  & \textit{ATE} &\textit{S.D.}  & \textit{ATE} &\textit{S.D.}   & \textit{ATE} &\textit{S.D.}  &  \textit{ATE} &\textit{S.D.}   & \textit{ATE} &\textit{S.D.}  & \textit{ATE} &\textit{S.D.} & \textit{ATE} &\textit{S.D.}\\
ORB-SLAM3\cite{orbslam3} &\xmark & 5.8 & 2.8  & 17.7 & 8.6 & 70.7 & 32.6  & 77.9 & 43.8 & \textbf{3.1} &1.6 &\textbf{3.5} &1.5 & 29.8 &15.2\\
Droid-VO\cite{teed2021droid} &\cmark & 5.4 & -  & 4.6 & -  &21.34 & -  & 46.0 & - &8.9 &- &5.9 &- &15.4 &-\\
DynaSLAM\cite{dynaslam} &\xmark & \textbf{3.0} & -  & \textbf{2.9} & - & \textbf{6.1} & -  &\textbf{7.8} & - & 4.9 &-& 3.9 &- & \textbf{4.8} &-\\
ReFusion\cite{refusion} &\cmark & 17.5 & -  & 25.4 & -  &28.9 & -  & 46.3 & - &30.2 & - & 17.9 &- & 27.7 &-\\
\midrule
\textit{NeRF based SLAM methods} & \textit{T/F} & \textit{ATE} &\textit{S.D.}  & \textit{ATE} &\textit{S.D.}   & \textit{ATE} &\textit{S.D.}  &  \textit{ATE} &\textit{S.D.}   & \textit{ATE} &\textit{S.D.}  & \textit{ATE} &\textit{S.D.} & \textit{ATE} &\textit{S.D.}\\
iMAP*\cite{imap} &\cmark &14.9  & 5.4  & 67.0  & 19.2  &28.3  & 12.9 & 52.8  & 20.9 &24.8  &11.2  &28.3  & 35.3 & 36.1 &17.5 \\
NICE-SLAM\cite{nice_slam} &\cmark &X  & X &66.8 & 20.0  &54.9  & 27.5 & 45.3 &17.5 & 21.2  & 13.1 &31.9  &13.6 &44.1 &18.4\\
Vox-Fusion\cite{voxfusion} &\cmark &65.7  & 30.9  & 82.1 & 52.0 &128.6  & 52.5 & 162.2 & 46.2  &43.9  & 16.5 & 47.5  & 19.5 &88.4 &36.3\\
Co-SLAM\cite{co-slam} &\cmark &28.8 & 9.6  & 20.6 & 8.1 &61.0 & 22.2  & 59.1 & 24.0 & 38.3 & 17.4 & 70.0 & 25.5 &46.3 & 17.8\\
ESLAM\cite{eslam} &\cmark &22.6 & 12.2  & 36.2 & 19.9 &48.0 & 18.7  & 51.4 & 23.2  &\textbf{12.4} & 6.6 & 17.7 & 7.5 & 31.4 & 14.7\\
RoDyn-SLAM(Ours) &\cmark & \textbf{7.9} & \textbf{2.7} &\textbf{11.5} &\textbf{6.1}  & \textbf{14.5}  & \textbf{4.6} & \textbf{13.8} & \textbf{3.5 } & 13.3 & \textbf{4.7} & \textbf{12.6}  & \textbf{4.7} &\textbf{12.3} &\textbf{4.4}\\
\bottomrule 
\end{tabular}
\label{table:bonn}
\end{center}
\end{table*}
\begin{figure}[t]
\scriptsize
\vspace{-2mm} 
\raggedleft
\includegraphics[width=0.47\textwidth]{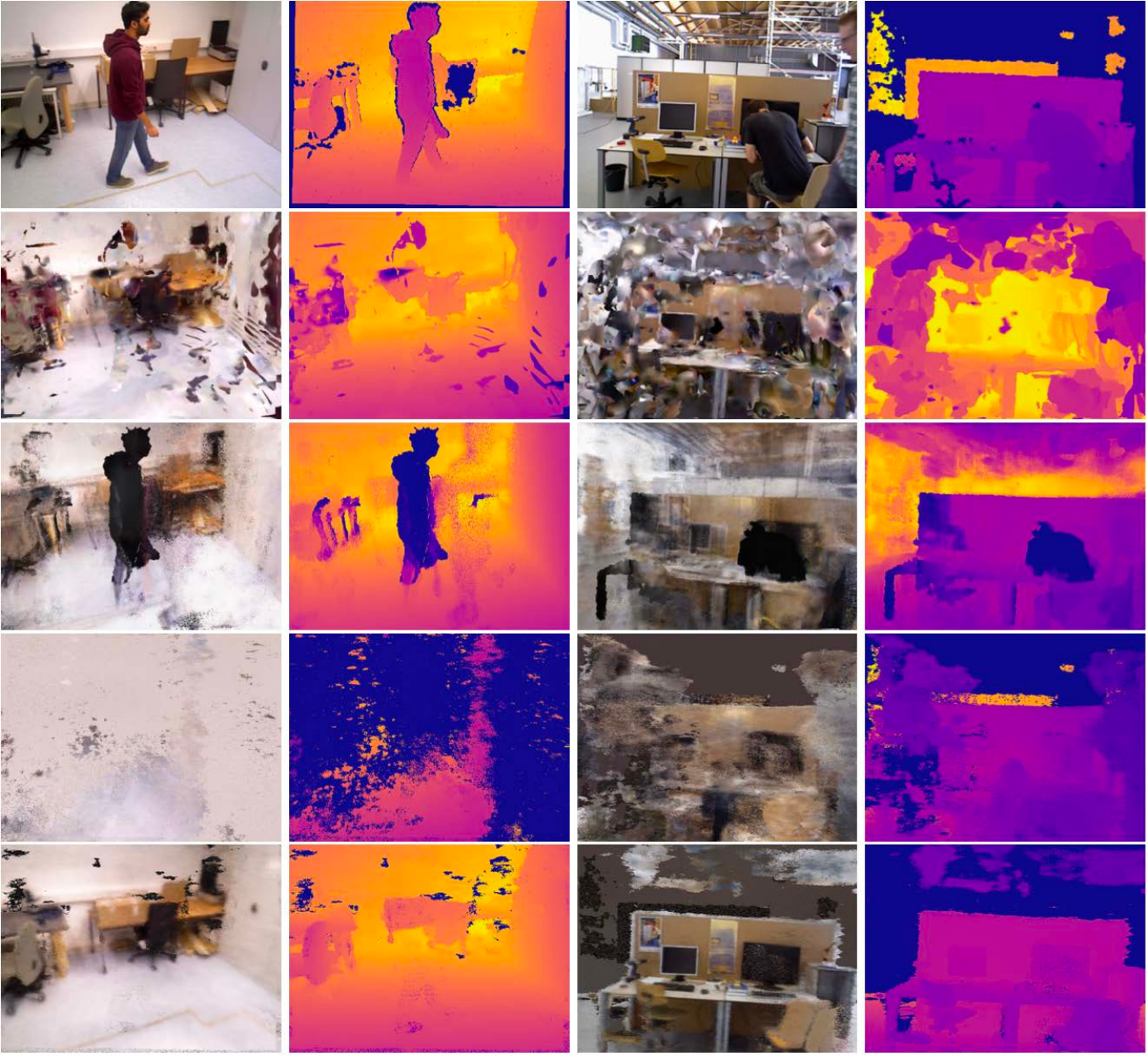}
\put(-218,-5){\scriptsize{\texttt{BONN person\_track} } }
\put(-92,-5){\scriptsize{\texttt{TUM f3\_walk\_xyz} }}
\put(-250,188){\scriptsize{\rotatebox{90}{Input GT}}}
\put(-250,137){\scriptsize{\rotatebox{90}{NICE-SLAM}}}
\put(-250,100){\scriptsize{\rotatebox{90}{ESLAM}}}
\put(-250,52){\scriptsize{\rotatebox{90}{Co-SLAM}}}
\put(-250,16){\scriptsize{\rotatebox{90}{Ours}}}
\vspace{-1mm}
\caption{\textbf{Visual comparison of the rendering image on the \textit{TUM} and \textit{BONN} datasets.}
}
\label{fig:vis_compare_render}
\vspace{-1mm}
\end{figure}
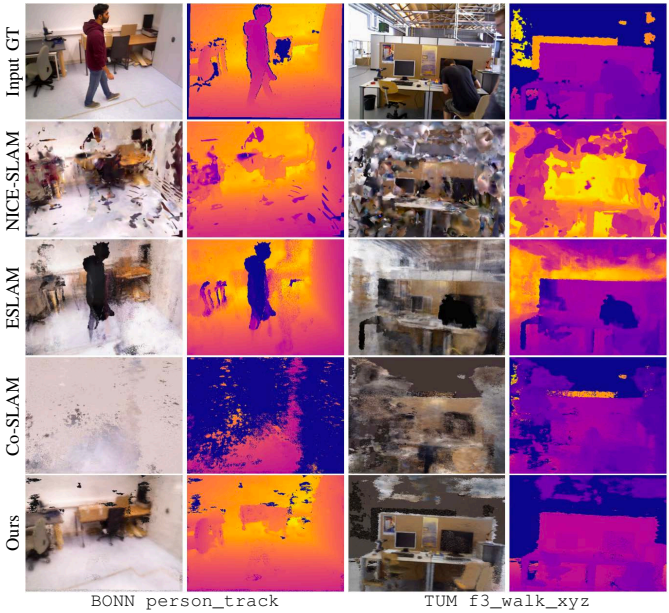

\subsection{Ablation study}
To demonstrate the effectiveness of the proposed methods in our system, we perform the ablation studies on seven representative sequences of the \textit{BONN} dataset, including \texttt{person\_tracking, balloon, balloon\_track, move\_no\_box}. As the semantic prior in the TUM dataset already covers most of the motion categories, we did not conduct ablation studies on this dataset. 
We compute the average ATE and STD results to show how different methods affect the overall system performance.
The results presented in Tab.~\ref{tab:ablation_study} demonstrate that all the proposed methods are effective in camera tracking. 
This suggests that fusing the optical flow mask and semantic motion mask can promote better pose estimation. 
At the same time, leveraging a divide-and-conquer pose optimization can effectively improve the robustness and accuracy of camera tracking. 

\subsection{Time consumption analysis}
As shown in Tab.~\ref{tab:time_consum}, we report time consumption (per frame) of the tracking and mapping without computing semantic segmentation and optical flow. Note that we pay more attention to evaluating the impact of our proposed methods on the baseline SLAM system's runtime. 
All the results were obtained using an experimental configuration of sampled 1024 pixels and 20 iterations for tracking and 2048 pixels and 40 iterations for mapping, with an RTX 3090 GPU in our laboratory server. Despite incorporating additional methods for handling dynamic objects, our system maintains a comparable level of computational cost to that of Co-SLAM. We also evaluate the time efficiency of our used optical flow and semantic segmentation network in our laboratory server, which required 97ms and 163ms respectively to process a single frame. Since semantic segmentation results can be pre-generated, the overall execution time of our optical flow fusion module is approximately 247ms. Note that Rodyn-SLAM is not optimized for real-time operation. With ongoing advancements in these research fields and improvements in computing power, the processing speeds for optical flow and semantic segmentation are expected to increase, ensuring they do not become bottlenecks for our method.

\setlength{\tabcolsep}{3pt}
\begin{table}[!t]
\vspace{-3mm}
\caption{Ablation study of the proposed method in our systems.}
\vspace{-4mm}
    \centering
    \footnotesize
    \renewcommand\arraystretch{1.2}
\centering
\begin{center}
        \begin{tabular}{c|c|c|c|c}
        
            \toprule
              & w/o Seg & w/o Flow & w/o Edge  & RoDyn-SLAM \\ \hline
            ATE RMSE (m) $\downarrow$&  0.3089 &   0.1793  &  0.2056 & \textbf{0.1354}  \\
            STD (m) $\downarrow$  &  0.1160 & 0.0739 & 0.0829  &  \textbf{0.0543} \\
            \bottomrule
        \end{tabular}
\label{tab:ablation_study}
 \end{center}
\end{table}

\setlength{\tabcolsep}{3pt}
\begin{table}[!t]
\vspace{-3mm}
\caption{Time comparison of different methods in our systems.}
\vspace{-4mm}
    \centering
    \footnotesize
    \renewcommand\arraystretch{1.2}
\centering
\begin{center}
        \begin{tabular}{c|c|c|c|c}
            \toprule
              & NICE-SLAM & ESLAM & Co-SLAM  & RoDyn-SLAM \\ \hline
            Tracking (ms) $\downarrow$&  3535.67 &   1002.52  &  174.47 & \textbf{159.06}  \\
            Mapping (ms) $\downarrow$ &  3055.58 & 703.69 & \textbf{565.50}  &  675.08 \\
            \bottomrule
        \end{tabular}
\vspace{-0.2in}
\label{tab:time_consum}
 \end{center}
\end{table}

\subsection{Visualization of Rendering Static Implicit Map}
To further demonstrate the performance of static scene reconstruction, we compared the rendered image with the ground truth pose obtained from the generated static implicit map. We selected two challenging sequences, \texttt{person\_track} from the BONN dataset and \texttt{f3\_walk\_xyz }from the TUM RGB-D dataset. As shown in Fig.~\ref{fig:vis_compare_render}, our method achieves a favorable rendering performance while enjoying the benefits of the proposed methods. Meanwhile, our methods can fill the hole which can not be captured in the original depth image. It can make the scene representation smoother and more complementing. We observed variations in rendering capabilities among different methods, which resulted in differences in the presentation quality. Note that our methods can be incrementally implemented in any existing baseline methods. Therefore, we don't focus on the actual performance of the code base Co-SLAM~\cite{co-slam} but solely on the proposed methods's ability and effectiveness in addressing dynamic scene challenges.

\section{Conclusion}
We present RoDyn-SLAM, a novel dense RGB-D SLAM with neural implicit representation for dynamic environments. 
The proposed system is able to estimate camera poses and recover 3D geometry in this challenging setup thanks to the motion mask generation that successfully filters out dynamic regions. 
To further improve the stability and robustness of pose optimization, a divide-and-conquer pose optimization algorithm is designed to enhance the geometry consistency between keyframe and non-keyframe with the edge warp loss. The experiment results demonstrate that RoDyn-SLAM achieves state-of-the-art performance among recent neural RGB-D methods in both accuracy and robustness. 
In future work, a more robust keyframe management method is a promising direction to improve the system further.

\bibliographystyle{IEEEtran}
\bibliography{IEEE-Transactions/mingo, IEEE-Transactions/references}

\end{document}